\documentclass[acmsmall,screen]{acmart}
\usepackage{algorithm}
\usepackage{algorithmic}
\usepackage{color}
\usepackage{multirow}
\usepackage{pifont}
\usepackage{makecell}
\usepackage{tabularx}
\usepackage{graphicx} % 用于插入图片
\usepackage{subcaption} % 用于创建子图

\AtBeginDocument{%
  }
\setcopyright{acmlicensed}
\copyrightyear{2026}
\acmYear{2026}
\acmDOI{XXXXXXX.XXXXXXX}

\copyrightyear{2026}
\acmYear{2026}
\acmConference[XXX '26]{XXXXXXXX}{XXXXXX}{XXXXXX}
\acmDOI{xxxxxx.xxxxxx}
\begin{document}

\title{GPU Temperature Simulation-Based Testing for In-Vehicle Deep Learning Frameworks}
%\title{ThermalGuardian: Temperature-Aware Testing of Automotive Deep Learning Frameworks}

\author{Yinglong Zou}
\affiliation{%
  \institution{State Key Laboratory for Novel Software Technology\\ Nanjing University}
  \country{China}
}
\email{652023320004@smail.nju.edu.cn}
\orcid{0009-0006-9375-7417}

\author{Juan Zhai}
\affiliation{%
  \institution{University of Massachusetts Amherst 
  \country{United States}}
}
\email{juanzhai@umass.edu}
\orcid{0000-0001-5017-8016}

\author{Chunrong Fang}
\authornote{corresponding author}
\affiliation{%
  \institution{State Key Laboratory for Novel Software Technology\\ Nanjing University}
  \country{China}
}
\email{fangchunrong@nju.edu.cn}
\orcid{0000-0002-9930-7111}

\author{Zhenyu Chen}
\affiliation{%
  \institution{State Key Laboratory for Novel Software Technology\\ Nanjing University}
  \country{China}
}
\email{zychen@nju.edu.cn}
\orcid{0000-0002-9592-7022}



\begin{abstract}
Deep learning models play a vital role in autonomous driving systems, supporting critical functions such as environmental perception. To accelerate model inference, these deep learning models' deployment relies on automotive deep learning frameworks, for example, PaddleInference in Apollo and TensorRT in AutoWare. However, unlike deploying deep learning models on the cloud, vehicular environments experience extreme ambient temperatures varying from -40°C to 50°C, significantly impacting GPU temperature. Additionally, heats generated when computing further lead to the GPU temperature increase. These temperature fluctuations lead to dynamic GPU frequency adjustments through mechanisms such as DVFS. However, automotive deep learning frameworks are designed without considering the impact of temperature-induced frequency variations. When deployed on temperature-varying GPUs, these frameworks suffer critical quality issues: compute-intensive operators face delays or errors, high/mixed-precision operators suffer from precision errors, and time-series operators suffer from synchronization issues. The above quality issues cannot be detected by existing deep learning framework testing methods because they ignore temperature's effect on the deep learning framework quality. To bridge this gap, we propose ThermalGuardian, the first automotive deep learning framework testing method under temperature-varying environments. Specifically, ThermalGuardian generates test input models using model mutation rules targeting temperature-sensitive operators, simulates GPU temperature fluctuations based on Newton's law of cooling, and controls GPU frequency based on real-time GPU temperature. Evaluated on PaddleInference and TensorRT, ThermalGuardian successfully detects 18 crashes and 3 NaN bugs, outperforming all baselines. Additionally, ThermalGuardian is 7 times faster than the best existing method, and achieves an 85\% operator coverage and a 100\% temperature-sensitive operator coverage, surpassing all baselines.
\end{abstract}

\begin{CCSXML}
<ccs2012>
 <concept>
  <concept_id>00000000.0000000.0000000</concept_id>
  <concept_desc>Do Not Use This Code, Generate the Correct Terms for Your Paper</concept_desc>
  <concept_significance>500</concept_significance>
 </concept>
 <concept>
  <concept_id>00000000.00000000.00000000</concept_id>
  <concept_desc>Do Not Use This Code, Generate the Correct Terms for Your Paper</concept_desc>
  <concept_significance>300</concept_significance>
 </concept>
 <concept>
  <concept_id>00000000.00000000.00000000</concept_id>
  <concept_desc>Do Not Use This Code, Generate the Correct Terms for Your Paper</concept_desc>
  <concept_significance>100</concept_significance>
 </concept>
 <concept>
  <concept_id>00000000.00000000.00000000</concept_id>
  <concept_desc>Do Not Use This Code, Generate the Correct Terms for Your Paper</concept_desc>
  <concept_significance>100</concept_significance>
 </concept>
</ccs2012>
\end{CCSXML}

\ccsdesc[500]{Software Engineering~Software Testing}

\keywords{Deep learning, Framework testing, Autonomous driving system}


\maketitle

\section{Introduction}
\label{introduction}

Deep learning (DL) techniques have become increasingly important in autonomous driving systems, playing a central role in critical tasks such as environmental perception \cite{ADSPerception1,ADSPerception2} (e.g., object detection \cite{object_detection,objectdetection_1,objectdetection_2} using camera images and LiDAR point clouds). DL models are widely deployed in industry-leading autonomous driving systems like Apollo \cite{Apollo} and AutoWare \cite{autoware}, where these models are able to handle complex, dynamic road scenarios. The widespread deployment of automotive DL models necessitates automotive DL frameworks, such as PaddleInference \cite{PaddleInference} in Apollo and TensorRT \cite{tensorrt} in AutoWare. These automotive DL frameworks provide efficient functional APIs for model inference, enabling hardware acceleration and optimized resource utilization essential for vehicles. However, compared with DL frameworks deploying DL models on the cloud, the deployment environment for automotive DL frameworks is more unstable. Specifically, vehicles operate in diverse geographical and climatic conditions where ambient temperatures may vary from -40°C in Arctic regions to 50°C in desert environments. Ambient temperatures directly impact GPU temperatures. Additionally, GPUs generate varying amounts of heat depending on computational intensity \cite{gpu_heat_generation}. Despite manufacturers implementing various temperature management solutions (e.g., liquid cooling \cite{liquid_cooling} and heat sink \cite{heat_sink}), vehicular hardware faces greater heat dissipation challenges than indoor server racks due to space constraints and limited airflow \cite{vehicle_heat_dissipation_challenge_1,vehicle_heat_dissipation_challenge_2}. When heat dissipation fails to keep pace with heat generation, GPU temperatures rise rapidly and may exceed the maximum threshold. These factors expose in-vehicle GPUs to extreme temperature fluctuations (e.g., ranging from -40°C in the polar environment to 90°C during sustained high-intensity computations). 

To avoid becoming too hot or too cold, GPU dynamically adjusts its frequency through mechanisms like Dynamic Voltage and Frequency Scaling (DVFS) \cite{DVFS1} based on its current temperature. When temperatures exceed the optimal temperature, GPUs reduce frequency to minimize heat generation; conversely, when temperatures drop below optimal temperature, frequency increases to boost heat generation. Unfortunately, automotive DL frameworks are designed without accounting for the impact of these temperature-induced frequency variations. When deployed in environments with varying temperatures, these frameworks experience critical quality issues. This has been confirmed by existing empirical studies \cite{temperature_related_bug}, for example, researchers find that high ambient temperatures can significantly decrease DL framework's inference throughput and further lead to inference delays and  precision drops. To make matters worse, some operators in DL frameworks are sensitive to temperature-induced GPU frequency variations. These operators are referred to as temperature-sensitive operators. When deployed in temperature-varying environments, these temperature-sensitive operators may experience severe quality issues.
Firstly, frequency reduction decreases floating-point operations per second (FLOPs), causing errors in compute-intensive operators such as GEMM convolution, where insufficient FLOPs prevent iterative computations from converging and ultimately result in precision errors or incorrect results. Secondly, lower frequency extends floating-point unit operation cycles. For high/mixed-precision operators (e.g., fp32 requiring 23-bit mantissa accumulation), incomplete carry-chain propagation within limited clock cycles leads to mantissa truncation errors and precision loss. Thirdly, frequency fluctuations cause clock jitter that disrupts synchronization in time-series operators (e.g., LSTM [16] and GRU [12]), leading to failures in state updating, which corrupts the temporal continuity of the hidden state and results in invalid or catastrophically erroneous outputs.

Unfortunately, the above quality issues cannot be successfully detected by existing DL framework testing methods \cite{lemon,muffin,gandalf} because these methods ignore the influence of the temperature on the DL framework quality. To bridge this gap, we propose ThermalGuardian, the first automotive DL framework testing method under simulated GPU temperature-varying environment. ThermalGuardian introduces two key innovations: it simulates GPU temperature varying environments and designs model mutation rules targeting temperature-sensitive operators to generate test input models that effectively reveal DL framework bugs under temperature fluctuations. Specifically, ThermalGuardian begins by collecting already-deployed DL models from autonomous driving systems (e.g., Apollo \cite{Apollo} and Autoware \cite{autoware}) to set up a seed model pool. It then loads test input tensors from widely adopted autonomous driving datasets such as KITTI \cite{KITTI} and NuScenes \cite{nuscenes}. After that, ThermalGuardian heuristically selects seed models from the seed model pool based on models' bug detection performance. Subsequently, ThermalGuardian heuristically selects and adopts model mutation rules to generate new test input models. These rules are designed focusing on temperature-sensitive operator types including compute-intensive, high/mixed-precision, and time-series operators. To emulate real-world deployment environments, ThermalGuardian periodically simulates GPU temperature fluctuations using Newton's law of cooling \cite{Newton_cooling_law} and dynamically adjusts GPU frequency according to the temperature. Within this simulated environment, differential testing is conducted across multiple automotive DL frameworks to detect quality issues including crashes, NaN bugs, and heavy inconsistencies. After testing, models with good bug detection performance are incorporated back into the seed model pool. Heuristic indicators are also updated to guide future iterations. In evaluation, we apply ThermalGuardian to test two widely-used automotive DL frameworks, including PaddleInference \cite{PaddleInference} in Apollo \cite{Apollo} and TensorRT \cite{tensorrt} in AutoWare \cite{autoware}. Experimental results show that ThermalGuardian detects 18 crashes and 3 NaN bugs, exceeding all existing methods. All detected bugs are reported to open-source communities, with 15 (12 crashes and 3 NaN bugs) confirmed and others awaiting response. These detected bugs contain 11 temperature-related bugs arising from Latency, Clock-Related Error, NaN, and Out-Of-Memory, and 10 non-temperature-related bugs arising from Unsupported Function, Tensor Shape Error, and Other. Additionally, ThermalGuardian is 7 times faster than the best existing method while achieving higher coverage (85\% operator coverage and 100\% temperature-sensitive operator coverage), demonstrating its efficiency and sufficiency.
% 温度的变化会影响GPU的频率，进而影响automotive DL框架的质量。Specifically，为了防止GPU温度过高或者过低，GPU会通过DVFS等机制根据温度实时地调节GPU的频率。当温度高于optimal 温度时，GPU会降低频率以减少热量产生；当温度低于optimal温度时，GPU会升高频率增加热量产生。GPU频率的变化会带来automotive DL框架的质量问题, 具体表现为如下三个方面: 1) GPU频率的降低会降低单位时间可完成的浮点运算量，进而导致进行高密度浮点运算的算子（e.g., GEMM convolution）的延迟或者错误。2） GPU频率的降低延长整个浮点运算单元的操作时间。For 高精度/混合精度的operators，整个进位链较长（e.g., fp32需要23位尾数累加）。当指定时钟周期内无法完成全部进位时，可能发生尾数截断等错误，造成精度损失。3）GPU频率的变化会造成时钟信号周期抖动，破坏时序算子（例如LSTM，GRU）的同步性，进而导致计算错误。不幸的是，现有的框架测试方法没能考虑到温度对框架质量的影响，simply 把模型部署到云端温度相对稳定的环境中。To bridge this gap，we propose ThermalGuardian，第一个automotive DL framework testing method 基于GPU温度不断变化的模拟环境。

Our main contributions are as follows:
\begin{itemize}
\item We propose the first temperature-aware automotive DL framework testing method. In this method, we simulate a temperature-varying environment, which is based on Newton's law of cooling and incorporates a temperature-based GPU frequency control.
\item We propose eight model mutation rules for generating test input models with temperature-sensitive operators, including compute-intensive, high-precision or mixed-precision, and time-series operators.
\item We implement the method as a tool named ThermalGuardian and apply ThermalGuardian to detect PaddleInference and TensorRT. Experimental results show that ThermalGuardian detects 18 crashes and 3 NaN bugs (with 15 confirmed by community developers), surpassing all baselines. Additionally, ThermalGuardian achieves a sevenfold speedup over the best baseline, an 85\% operator coverage, and a 100\% temperature-sensitive operator coverage.
\end{itemize}

We make all of our data and source code publicly available on Github: \url{https://github.com/ThermalGuardian/ThermalGuardian}.
\section{Background}
\label{background}

\subsection{Deep Learning in Autonomous Driving Systems}

Deep learning (DL) is fundamental to modern autonomous driving systems, enabling perception \cite{ADSPerception1,ADSPerception2} and decision-making \cite{decision_making} capabilities. These systems use DL models to process multi-modal sensor data (e.g., camera images, LiDAR point clouds) in real-time, interpreting complex driving environments \cite{dlADS_overview1,dlADS_overview2}. DL models depend on underlying automotive DL frameworks (e.g., TensorRT \cite{tensorrt} and PaddleInference \cite{PaddleInference}). Automotive DL frameworks are pivotal for autonoumous driving systems' efficiency, safety, and robustness, managing hardware resource scheduling, memory optimization, and real-time inference execution. To ensure the quality and reliability of these frameworks, in this work, we design the first automotive DL framework testing method.

\subsection{Temperature Fluctuations in Vehicular Environments}

Autonomous driving systems operate in extreme environments with ambient temperatures ranging from -40°C in polar regions to 50°C in desert regions \cite{polar_temperature,desert_temperature}. These ambient temperature fluctuations directly influence the GPU temperature through heat transfer mechanisms \cite{gpu_cooling}. Simultaneously, the substantial heat generated by computational workloads—especially during DL inference for perception tasks—further elevates the GPU temperature \cite{gpu_heat_generation}.
The combination of widely fluctuating ambient temperature and variable heat generation leads to significant and dynamic fluctuations in GPU temperature \cite{gpu_cooling}. To avoid GPU temperature from overheating or undercooling, GPUs employ dynamic voltage and frequency scaling (DVFS) \cite{DVFS1}. This technology automatically adjusts GPU frequency based on real-time GPU temperature. Specifically, when GPU temperature approach the upper safety limit, DVFS reduces the frequency to decrease power dissipation and prevent overheating, while near the lower threshold, frequency is increased to prevent undercooling \cite{DVFS2,DVFS3}. This frequency introduces potential latency increases in convolutional operators, numerical precision issues in floating-point calculations, and synchronization challenges in time-series operators. To ensure automotive DL framework quality in GPU temperature-varying environments, ThermalGuardian simulates GPU temperature fluctuations and controls GPU frequency according to real-time GPU temperature.
\section{Methodology}
\label{methodology}

\subsection{Overview}
\label{overview}

\begin{figure*}[htpb]
    \centering
    \includegraphics[width=0.95\textwidth]{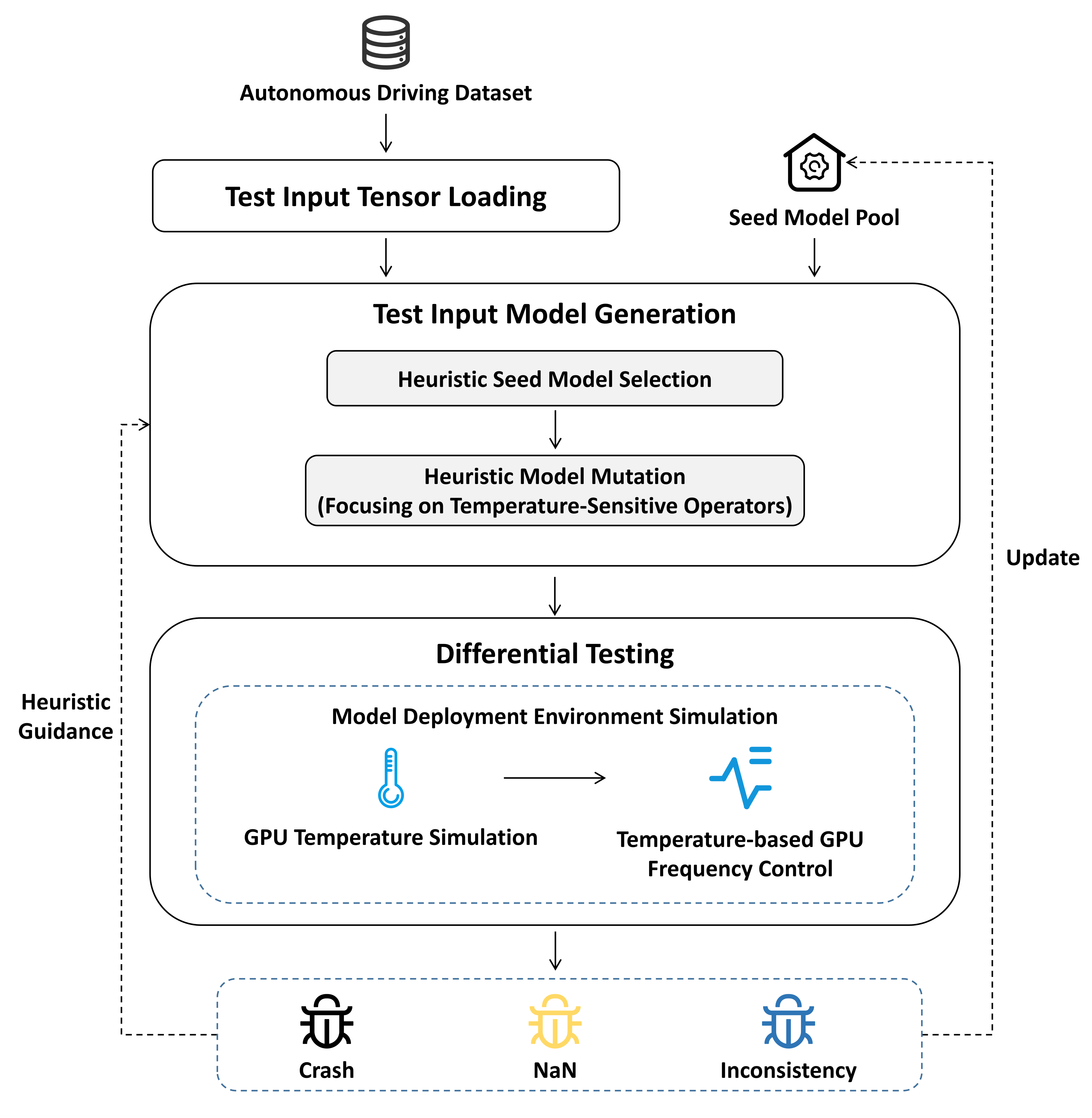}
    \Description{workflow}
    \caption{The Workflow of ThermalGuardian}
    \label{workflow}
\end{figure*}

To ensure the quality of in-vehicle DL frameworks under temperature-varying environments, we propose ThermalGuardian, an automotive DL framework testing approach based on GPU temperature simulation. ThermalGuardian designs eight model mutation rules focusing on three types of temperature-sensitive operators to generate test input models, and simulates temperature-varying deployment environments to conduct differential testing. In simulation, ThermalGuardian periodically emulates GPU temperature fluctuations based on Newton's law of cooling \cite{Newton_cooling_law} and adjusts GPU frequency based on real-time GPU temperature. Figure \ref{workflow} shows the overall workflow of ThermalGuardian. Specifically, before testing, all DL models already-deployed in tested autonomous driving systems (including Apollo and Autoware) are collected as seed models in the seed model pool. ThermalGuardian then loads test input tensors from widely-used autonomous driving datasets (e.g., KITTI \cite{KITTI}, NuScenes \cite{nuscenes}), heuristically selects seed models from the pool based on their bug detection performance, and applies eight model mutation rules targeting temperature-sensitive operators to generate new test input models. To simulate deployment environments for models generated above, ThermalGuardian designs two modules: \ding{192} periodic GPU temperature simulation via Newton's law of cooling, and \ding{193} real-time GPU frequency control based on GPU temperature. Under simulated environment, ThermalGuardian executes differential testing across different autonomous driving systems to detect automotive DL framework bugs. After differential testing, models demonstrating excellent bug detection performance are incorporated into the seed model pool. Additionally, heuristic indicators based on model's bug detection performance are updated to guide subsequent iterations of test input model generation.

\subsection{Test Input Tensor Loading}
\label{test input tensor loading}

Tensors serve as a critical part of test inputs, which are loaded from autonomous driving datasets. For better representativeness, ThermalGuardian adopts two widely used datasets, including KITTI \cite{KITTI} and NuScenes \cite{nuscenes}. However, raw data from these datasets cannot be directly loaded as tensors due to incompatibility with automotive model requirements. Specifically, intelligent vehicles primarily perceive the surroundings through cameras and LiDAR. Accordingly, there are two data modalities in autonomous driving datasets: camera images and LiDAR point clouds. For camera images, since models deployed on vehicles must interface with camera hardware, these models can only process images whose size are specified by camera configuration. To meet this requirement, ThermalGuardian scales and crops camera images to change their size according to the camera configuration. For LiDAR point clouds, due to the sparse distribution of points, automotive models struggle to capture features directly. To better extract features, ThermalGuardian adopts voxelization to convert LiDAR point clouds into voxels---a well-organized data structure capable of resolving the sparsity of LiDAR cloud points  \cite{voxelization}.

\subsection{Test Input Model Generation}
\label{test input model generation}

DL models are another important part of test inputs, which invoke operators provided by DL frameworks. To effectively test automotive DL frameworks, we design a novel test input model generation method focusing on temperature-sensitive operators. This method addresses a critical limitation in existing DL framework testing methods \cite{lemon}, which fail to consider the impact of temperature on DL framework quality. The types of temperature-sensitive operators and the model mutation rules designed based on these operators will be introduced in Section \ref{model mutation rule design}. ThermalGuardian generates new models by designing model mutation rules and applying these rules to seed models. The process begins with the establishment of a seed model pool containing already-deployed models. Subsequently, ThermalGuardian heuristically selects seed models (Section \ref{heuristic seed model selection}) and model mutation rules (Section \ref{heuristic model mutation rule selection}). These seed models and rules are then adopted to generate new test input models. Following differential testing based on the generated models, ThermalGuardian updates the heuristic indicators and incorporates newly generated models with strong bug detection performance into the seed model pool (Section \ref{bug detection}).

\subsubsection{Heuristic Seed Model Selection}
\label{heuristic seed model selection}

Seed models are essential input for mutation-based model generation methods, which are selected in two steps. 

\textbf{1) Set up and maintain a seed model pool.}
To ensure the generated DL models align closely with real-world applications, we collect models already-deployed in autonomous driving systems as seed models. Specifically, we select the most widely-used autonomous driving systems (including Apollo and Autoware). In these systems, DL models are overwhelmingly deployed in the perception module. Consequently, we select this representative module and collect all models already-deployed in the perception module as seed models in the pool. The seed model pool is maintained throughout testing. ThermalGuardian updates it each iteration by adding newly generated models that demonstrate good bug detection performance.

\textbf{2) Heuristically select seed models in the pool.} To generate test input models more efficiently, $performance$ quantitatively reflects each seed model's bug detection performance. ThermalGuardian's definition for $performance$ is as follows.
$$ performance = 
\left\{
\begin{array}{ll}
m_d & \text{(when crashes or NaN bugs are triggered)}\\
MAE & \text{(when no bugs are triggered)}
\end{array}
\right.
$$
Specifically, when triggering crashes and NaN bugs, $performance$ is $m_d$, which is the mean value of the test input tensor. When no bugs are detected, ThermalGuardian focuses on the inconsistency between DL frameworks, setting $performance$ to the Mean Absolute Error (MAE) among output tensors (will introduce in Section \ref{bug detection}). The $performance$ for newly generated models is calculated and recorded after each round of differential testing (Section \ref{differential testing}). The $performance$ for each initial seed model is calculated based on bugs already triggered in the original autonomous driving system.

Based on the model's $performance$, ThermalGuardian heuristically selects models with good bug detection performance. Specifically, to avoid consistently choosing the same seed model and potentially converging to a local optimum, ThermalGuardian first randomly draws 10\% of the models from the seed model pool to form a candidate subset. From this subset, the model with the highest $performance$ is then chosen as the seed model for mutation in the current iteration.

% The detailed procedure is as follows. Step 1) ThermalGuardian randomly selects k groups of seed models from the seed model pool, with each group comprising 10\% of the total seed models. Step 2) Within each group, ThermalGuardian filters out the top-k seed models with highest $performance$. Step 3) Among the $k \times k$ seed models selected from all groups, ThermalGuardian further filters the top-k seed models with highest $performance$. To accelerate the iteration speed, k is set to 1 in ThermalGuardian.

\subsubsection{Model Mutation Rule}
\label{model mutation rule design}

As introduced in Section \ref{test input model generation}, existing methods fail to consider temperature's influence on automotive DL framework quality. To bridge this gap, we design eight model mutation rules focusing on three types of temperature-sensitive operators (including compute-intensive operators, high-precision/mixed-precision operators, and time-series operators). The purpose of these rules is to generate test input models with more temperature-sensitive operators, and thus triggering more temperature-related DL framework bugs. All model mutation rules are shown in Table \ref{table_model mutation rules in ThermalGuardian}. In this table, the first column represents targeted operator types. The last three columns represent the ID, name, and description of the mutation rule, respectively. \textbf{Mutation Rules 1--7} are designed for temperature-sensitive operators. Since the combination of non-temperature-sensitive operators also contributes to generating effective test input models, \textbf{Mutation Rule 8} is designed for inserting, deleting, and replacing non-temperature-sensitive operators.

\begin{table*}[htpb]\small
    \caption{Model Mutation Rules in ThermalGuardian}
    \label{table_model mutation rules in ThermalGuardian}
    \centering
    \begin{tabularx}{0.99\textwidth}{m{0.18\textwidth}m{0.005\textwidth}>{\centering}m{0.25\textwidth}m{0.45\textwidth}}
    \cline{1-4}
       \centering Related Operator & ID & Mutation Rule & Description \\ 
       \cline{1-4}
        \multirow{2}{0.18\textwidth}{\centering Compute-Intensive Operator} & 1 & GEMM Convolution Insertion & Inserts a convolution with GEMM optimization enabled. \\ 
        \cline{2-4} 
        & 2 & MatMul Product Insertion& Inserts the operator $matmul$. \\ 
        \cline{1-4}
        \multirow{5}{0.18\textwidth}{\centering High-Precision/\\Mix-Precision Operator} & 3 & High Precision Operator Replacement& Replaces an operator with a functionally equivalent high-precision operator (e.g., substitutes an int8 convolution with an fp32 convolution). \\ 
        \cline{2-4}
        & 4 & Mix Precision Operator Replacement& Replaces an operators with a functionally equivalent mixed-precision operator (e.g., substitutes a convolution operator with an int8/fp16 mixed-precision convolution). \\ 
        \cline{1-4}
        \multirow{5}{0.18\textwidth}{\centering Time-Series Operator} & 5 & Recurrent Neural Network (RNN) Insertion & Inserts a RNN (e.g., unidirectional RNN and bidirectional RNN). \\ 
        \cline{2-4}
        & 6 & Long Short-Term Memory (LSTM) Insertion& Inserts a LSTM operator (e.g., single-layer LSTM, multi-layer LSTM, and bidirectional LSTM). \\ 
        \cline{2-4}
        & 7 & Gated Recurrent Unit (GRU) Insertion & Inserts a GRU operator (e.g., single-layer GRU, multi-layer GRU, and bidirectional GRU). \\ 
        \cline{1-4}
        \centering Non-Temperature-Sensitive Operator & 8 & Random Operator Replacement (ROR) & Replaces an operator with a randomly selected non-temperature-sensitive operator. \\ 
        \cline{1-4}
    \end{tabularx}
\end{table*}

\textbf{Mutation Rules 1 \& 2.} These mutation rules focus on compute-intensive operators. Compute-intensive operators are those that perform high-density floating-point computations. These operators require numerous parallel computational units to operate continuously. The increase in GPU temperature triggers frequency reductions, thereby decreasing the volume of floating-point operations completed per unit time, potentially causing computational delays or errors. To test whether compute-intensive operators incur errors during GPU temperature fluctuations, we select the most commonly used compute-intensive operators (including GEMM convolution and MatMul product) to design mutation rules. Specifically, \textbf{Mutation Rule 1} targets GEMM convolution. In this mutation rule, ThermalGuardian inserts convolution operators with GEMM optimization enabled. The convolution type is randomly selected among three widely-used types: convolution, depthwise convolution, and separable convolution. \textbf{Mutation Rule 2} targets MatMul product; in this mutation rule, ThermalGuardian inserts the operator $matmul$ into the DL model.

\textbf{Mutation Rules 3 \& 4.} These mutation rules focus on high-precision/mixed-precision operators. High-precision operators refer to those supporting high computational precision such as fp32 \cite{high-precision_operator}. Mixed-precision operators refer to those flexibly supporting multiple precisions such as int8 and fp16 \cite{mix-precision_operator}. Due to the precision requirements of these operators, they are more sensitive to temperature variations. Specifically, temperature increases cause GPU frequency reductions, which decrease transistor switching speeds, potentially resulting in floating-point units failing to complete carry computations within designated clock cycles and thereby causing carry errors. Additionally, fp32 requires 23-bit mantissa bits to operate accumulation. Temperature-induced GPU frequency reductions prolong the entire carry chain operation time, potentially leading to mantissa truncation errors when 23-bit carries cannot be completed within the specified clock cycle, resulting in precision loss. To test whether temperature affects the quality of high-precision/mixed-precision operators, we design model mutation rules targeting these operators. \textbf{Mutation Rule 3} replaces operators in the model with functionally equivalent high-precision operators, for example, replacing int8 convolution with fp32 convolution. \textbf{Mutation Rule 4} replaces operators in the model with functionally equivalent mixed-precision operators, for example, replacing convolution with int8/fp16 mixed-precision convolution.

\textbf{Mutation Rules 5--7.} These mutation rules focus on time-series operators, which have strict computational order. Temperature-caused GPU frequency reductions extend computation time and bring clock period jitter, compromising synchronization and leading to computational errors. For example, LSTM gates such as the forget gate and input gate require precise temporal alignment, and clock jitter can disrupt the synchronization of gating signals, causing state update errors \cite{time-series_operator_1,time-series_operator_4}. To test whether temperature variations affect the quality of time-series operators, we design mutation rules targeting three widely used time-series operators including RNN, LSTM, and GRU \cite{time-series_operator_2,time-series_operator_3}. \textbf{Mutation Rule 5--7} insert RNN, LSTM, and GRU respectively. For representative, we adopt multiple data flow directions and multiple layers. Specifically, for RNN, we adopt both unidirectional RNN and bidirectional RNN. For LSTM, we adopt single-layer LSTM, multi-layer LSTM, and bidirectional LSTM. For GRU, we adopt single-layer GRU, multi-layer GRU, and bidirectional GRU.

\textbf{Mutation Rule 8.} This mutation rule is designed to generate models with various non-temperature-sensitive operators. It replaces an operator with another randomly selected non-temperature-sensitive operator. This mutation rule not only replaces operators in the model but also inserts and removes operators in the model. Specifically, the operator $None$ is an available candidate operator. When the operator before replacement is $None$ but the operator after replacement is not $None$, it means inserting an operator into the model. When the operator before replacement is not $None$ but the operator after replacement is $None$, it means removing an operator from the model. In other cases, it means replacing the operator in the model.

% In addition to the above model mutation rules, following existing methods \cite{lemon}, ThermalGuardian also supports randomly inserting, replacing the operator in the model.
% % todo：这里问问应该怎么改

Following existing methods \cite{gandalf,lemon}, the operator insertion and replacement in above mutation rules are implemented by graph modification. Specifically, ThermalGuardian adopts Directed Acyclic Graph (DAG) to represent each seed model. The DAG comprises a vertex set $V$ and an edge set $E$, in which vertices correspond to tensors and edges correspond to operators. Each edge has its own type, and distinct edge types denote different operators. In DAG, the insertion of new operators is represented as adding new edges and vertices, while operator replacement is represented as modifying edge types. The position of the insertion and replacement is randomly selected.

\subsubsection{Heuristic Model Mutation Rule Selection}
\label{heuristic model mutation rule selection}
ThermalGuardian generates a new model by selecting a model mutation rule (designed in Section \ref{model mutation rule design}) and adopting it on the selected seed model (see Section \ref{heuristic seed model selection} for seed model selection). To control the selection of model mutation rules and generate models triggering more bugs, we design a heuristic indicator called $contribution$. $contribution$ quantifies the cumulative contribution of each model mutation rule to bug detection. The initial value of $contribution$ is set to 0. $contribution$ is updated based on the bug detection performance of the seed model (called $performance_{seed}$) and the newly generated model (called $performance_{new}$). Specifically, ThermalGuardian updates the $contribution$ of selected model mutation rule after each iteration of differential testing according to the following formula:
$$contribution_i = contribution_{i-1} + \Delta performance$$
where $contribution_{i}$ and $contribution_{i-1}$ are the $contribution$ of selected model mutation rule after and before update. $\Delta performance = performance_{new} - performance_{seed}$, where $performance_{new}$ is the bug detection performance of the newly generated model, and $performance_{seed}$ is the bug detection performance of the selected seed model. Based on $contribution$, ThermalGuardian controls the probability $p$ of selecting each model mutation rule as follows: 
$$p = \frac{contribution}{\sum_{i=1}^{n}contribution_i}$$
where $n$ is the total number of available model mutation rules.

\subsection{Differential Testing}
\label{differential testing}

Based on test input tensors (loaded in Section \ref{test input tensor loading}) and test input models (generated in Section \ref{test input model generation}), ThermalGuardian adopts differential testing on widely used automotive DL frameworks. The overall process of differential testing is as follows. Firstly, ThermalGuardian simulates model deployment scenario with varying temperatures (see Section \ref{model deployment environment simulation}). The purpose of this simulation is to test whether temperature variations affect the quality of automotive DL frameworks. As introduced in Section \ref{introduction}, the GPU temperature is influenced by ambient temperature and heat dissipation. Due to mechanisms like DVFS \cite{DVFS1,DVFS2,DVFS3}, GPU frequency varies with temperature fluctuates, potentially leading to bugs such as precision degradation and time series disruption. In the above simulated environment, ThermalGuardian deploys the test input models (generated in Section \ref{test input model generation}) to different autonomous driving systems. Specifically, for representativeness, ThermalGuardian selects the two most widely-used autonomous driving systems, Apollo \cite{Apollo} and AutoWare \cite{autoware}, and follows their native model deployment workflows to test the integrated in-vehicle DL frameworks within these systems. For example, in Apollo, models are developed and trained on the DL framework PaddlePaddle \cite{paddlepaddle}, and subsequently deployed to the integrated in-vehicle DL framework PaddleInference \cite{PaddleInference} within the Apollo system; in AutoWare \cite{autoware}, models are developed and trained on the DL framework PyTorch \cite{pytorch}, and deployed to the integrated in-vehicle DL framework TensorRT \cite{tensorrt} within the AutoWare system. By inputting identical tensors (loaded in Section 3.2.1) and comparing the execution results of models between cloud-based DL frameworks (e.g., PaddlePaddle and PyTorch) and automotive DL frameworks (e.g., PaddleInference and TensorRT), ThermalGuardian detects bugs in automotive DL frameworks. Following existing methods \cite{muffin,gandalf}, ThermalGuardian focuses on crashes, NaN bugs, and heavy inconsistencies among DL frameworks (see Section \ref{bug detection}).   

\subsubsection{Model Deployment Environment Simulation}
\label{model deployment environment simulation}

The GPU temperature is influenced by ambient temperature and heat dissipation, thereby causing variations in GPU frequency. To simulate this environment, ThermalGuardian first emulates GPU temperature changes based on Newton's law of cooling \cite{Newton_cooling_law}, then controls GPU frequency according to the temperature. Details are as follows.

\textbf{1) GPU Temperature Simulation.}
\label{GPU temperature simulation}
ThermalGuardian simulates GPU temperature fluctuations based on Newton's law of cooling \cite{Newton_cooling_law}, which is a fundamental physical principle describing the temperature dynamics of objects. It states that the rate of heat dissipation is proportional to the temperature difference between the object and its environment. A larger temperature difference accelerates temperature changes, while a smaller difference slows them down. This law is particularly well-suited for simulating GPU temperature fluctuations because GPU generates heat during operation, creating a temperature difference with its surroundings. Newton's law of cooling formula is as follows:
$$T(t) = T_{env} + (T_{initial} - T_{env}) \times e^{-kt}$$
where $T(t)$ is the temperature of the object at time $t$, representing the instantaneous thermal state during cooling or heating. $T_{env}$ is the ambient temperature, the constant equilibrium temperature of the surroundings to which the object ultimately tends. $T_{initial}$  is the initial temperature of the object at $t = 0$, serving as the starting point for thermal evolution. $k$ is the cooling coefficient, a positive constant scaling the rate of heat dissipation (unit: $s^{-1}$), determined by material properties and surface geometry. $t$ is the time elapsed since cooling began (unit: seconds), defining the progression of thermal equilibrium. $e$ is the natural exponent.

In GPU's temperature fluctuation process, the cooling coefficient $k$ is determined by GPU's type. For GPUs with the same type, $k$ is a fixed constant (according to each GPU's official website \cite{NVIDIA_GPU_website}). By assigning distinct values to $T_{env}$ and $T_{initial}$, ThermalGuardian simulates various GPU temperature fluctuation scenarios. Specifically, for each type of GPU, the official documentation \cite{nvidia_document} specifies the maximum supported temperature threshold $T_{max}$, the minimum supported temperature threshold $T_{min}$, and the nominal temperature $T_{nominal}$ (the optimal operating temperature during normal GPU operation). Accordingly, based on Newton's law of cooling, ThermalGuardian emulates six common GPU temperature fluctuation scenarios, as shown in Table \ref{table_temperature_scenario}. In this table, the first column and the second column respectively represent the ID and description of each GPU temperature fluctuation scenario, while the subsequent two columns denote the value of $T_{initial}$ and $T_{env}$ within these scenarios. The first three scenarios in this table represent GPU heating up processes, while the last three represent GPU cooling down processes.

\begin{table}[htpb]
    \caption{The Value of $T_{env}$ and $T_{initial}$ in ThermalGuardian's Simulated Scenarios}
    \centering
    \begin{tabular}{cccc}
    \hline
         ID & Scenario Name & $T_{initial}$ Value & $T_{env}$ Value \\ \hline
         1 & Computing in Cold Environment with Heavy Workloads& $T_{min}$ & $T_{max}$ \\ 
        2 & Computing in Cold Environment with Nominal Workloads& $T_{min}$ & $T_{nominal}$ \\ 
         3 & Computing in Nominal Environment with Heavy Workloads& $T_{nominal}$ & $T_{max}$ \\ 
         4 & Cooling in Cold Environment after Heavy Workloads & $T_{max}$ & $T_{min}$ \\ 
         5 & Cooling in Cold Environment after Nominal Workloads & $T_{nominal}$ & $T_{min}$ \\ 
         6 & Cooling in Nominal Environment after Heavy Workloads & $T_{max}$ & $T_{nominal}$ \\ \hline
    \end{tabular}
    \label{table_temperature_scenario}
\end{table}

Specifically, ThermalGuardian simulates six temperature-varying scenarios as follows:
\begin{itemize}
    \item 
In \textbf{Scenario 1}, GPU performs intensive computation in a cold environment, which generates heats that elevate GPU temperature from $T_{min}$ to $T_{max}$. 
    \item 
In \textbf{Scenario 2}, GPU performs nominal computation in a cold environment, which generates heats that elevate GPU temperature from $T_{min}$ to $T_{nominal}$. 
    \item 
In \textbf{Scenario 3}, a nominally operating GPU temporarily performs intensive computation, generating heats that elevate GPU temperature from $T_{nominal}$ to $T_{max}$. 
    \item 
In \textbf{Scenario 4}, in a cold environment, after performing intensive computation, the GPU becomes temporarily idle, cooling from $T_{max}$ to $T_{min}$. 
    \item 
In \textbf{Scenario 5}, in a cold environment, after performing nominal computation, the GPU becomes temporarily idle, cooling from $T_{nominal}$ to $T_{min}$. 
    \item 
In \textbf{Scenario 6}, after performing intensive computation, GPU transitions to nominal computation, resulting in reduced heat generation, cooling the temperature from $T_{max}$ to $T_{nominal}$.
\end{itemize}

\textbf{2) Temperature-Based GPU Frequency Control.}
\label{temperature-based GPU frequency control}
As introduced in Section \ref{introduction}, to prevent bugs caused by excessively high or low GPU temperatures, GPUs dynamically adjust their frequency via mechanisms such as DVFS (Dynamic Voltage and Frequency Scaling) \cite{DVFS1,DVFS2,DVFS3} to regulate heat generation. Specifically, when temperature is lower than $T_{nominal}$, the GPU increases frequency to generate more heat. Conversely, when temperature is higher than $T_{nominal}$, GPU reduces frequency to minimize heat production. Following the DVFS mechanism, ThermalGuardian simulates the GPU frequency variations under GPU temperature fluctuations as follows.

$$
f(T) = f_{\text{base}} \times 
\left\{
\begin{array}{ll}
1 + \gamma \cdot \dfrac{T_{\text{nominal}} - T}{T_{\text{nominal}} - T_{\text{min}}} & T < T_{\text{nominal}} \quad \text{(low 
temperature frequency increase)} \\
\\[0.25em]  
1 - \alpha \cdot \dfrac{T - T_{\text{nominal}}}{T_{\text{max}} - T_{\text{nominal}}} & T \geq T_{\text{nominal}} \quad \text{(high temperature frequency decrease)}
\end{array}
\right.
$$

In this formula, $T$ denotes the real-time GPU temperature. $f(T)$ denotes the real-time GPU frequency. $T_{min}$, $T_{max}$, and $T_{nominal}$ $f_{base}$ denotes the same as Section \textit{GPU Temperature Simulation}. $f_{base} $ denotes the baseline GPU frequency under the nominal temperature. $\gamma$ denotes the low-temperature scaling coefficient, dictating the rate of frequency increase when $T < T_{nominal}$. $\alpha$ denotes the high-temperature throttling coefficient, governing the rate of frequency reduction when $T \geq T_{nominal}$. The values of $\alpha$ and $\gamma$ are determined by the GPU type. Taking the NVIDIA GeForce RTX 4090D as an example, according to the official website, $\alpha$ is 0.15 and $\gamma$ is 0.05.

% todo: alpha 和 mamma的取值由芯片设计决定，以RTX 4090D为例，实测数据显示，alpha为xxx，gamma为xxx。

\subsubsection{Bug Detection}
\label{bug detection}

Based on generated test input models and tensors, ThermalGuardian performs differential testing between cloud-end (e.g., PaddlePaddle and PyTorch) and automotive DL frameworks (e.g., PaddleInference and TensorRT) to detect automotive DL framework bugs , focusing on two types of bugs: crashes and NaN bugs. And additionally, ThermalGuardian also focuses on the heavy inconsistency among DL frameworks. Among them, crashes are detected when the same model executes successfully on cloud-end DL framework but breaks down on the automotive DL framework. To avoid duplication, when a crash occurs, ThermalGuardian automatically records the model's execution logs and computes the textual cosine similarity \cite{text_similarity} between the newly recorded log and the logs recorded in previous rounds. Cosine similarity measures the similarity between two strings, ranging from -1 to 1, where 1 indicates identical content, -1 indicates completely opposite content, and 0 indicates no correlation. Therefore, if the newly recorded execution logs are identical to the previously recorded execution logs (i.e., the textual cosine similarity equals to 1), it is considered that ThermalGuardian has repeatedly triggered the same bug which will not be counted again. NaN bugs are detected when an NaN orrurs in the output tensor on the automotive DL framework but not on the cloud-end DL framework. Heavy inconsistency is detected based on the Mean Absolute Error (MAE) among DL frameworks. The calculation for MAE is: $$\text{MAE(x,y)} = \frac{1}{n} \sum_{i=1}^{n} |x_i - y_i|$$

where $x$, $y$ represent the output tensors in different DL frameworks. $n$ represents the number of data points in an output tensor. $x_{i}$ and $y_{i}$ represent the $i$-th data point in each output tensor. Following existing methods \cite{predoo}, when MAE among DL frameworks is larger than 0.15, a heavy inconsistency is detected. To guide the following iterations, after differential testing, ThermalGuardian calcutes and records each newly generated model's  $performance$ (see Section \ref{heuristic seed model selection} for $performance$ calculation). Based on $performance$, ThermalGuardian updates the $contribution$ of selected model mutation rule (as introduced in Section \ref{heuristic model mutation rule selection}). Additionally, ThermalGuardian adds models triggering crashes, NaN bugs, and heavy inconsistencies into the seed model repository.
\section{Evaluation}
\label{evaluation}

To evaluate ThermalGuardian's effectiveness and efficiency, we deploy ThermalGuardian to test two widely-used automotive DL frameworks, PaddleInference in Apollo and TensorRT in AutoWare. We select LEMON \cite{lemon}, Muffin \cite{muffin}, and Gandalf \cite{gandalf} as our baselines. These methods are all state-of-the-art methods for testing cloud-end DL frameworks (e.g., TensorFlow \cite{tensorflow} and PyTorch \cite{pytorch}). Since there are no methods designed for testing automotive DL frameworks, these baselines are the most relative to our method.

\subsection{Experimental Setup}
\label{experimental setup}

We establish the test environment on an Ubuntu 18.04 workstation equipped with an NVIDIA GeForce RTX 4090D GPU and an Intel Core i7-12700KF processor. The NVIDIA driver 535.216.01 is installed along with CUDA 12.4 and CUDNN 9.6.0 to enable GPU acceleration. Apollo 9.0 is downloaded from its official repository, and its integrated PaddleInference v2.6.2 framework is deployed. For Autoware, TensorRT v8.5.3.1 is configured following standard installation guides. ThermalGuardian is integrated into these environments using Python 3.8, with simulation parameters set according to the NVIDIA official website \cite{NVIDIA_GPU_website}. Specifically, for the NVIDIA RTX 4090D, the cooling coefficient $k$ in Newton's law of cooling is set to 0.015. In the DVFS mechanism, the high-temperature throttling coefficient $\alpha$ is 0.15, and the low-temperature scaling coefficient $\gamma$ is 0.05. The nominal temperature $T_{nominal}$ is set to 40°C,  $T_{min}$ to -40°C, and $T_{max}$ to 90°C. All baselines are cloned from GitHub and configured with default parameters as specified in their respective papers. All generated test input models are converted to PaddlePaddle v2.6.2 and PyTorch v1.13.1 formats. Subsequently, ThermalGuardian deploys these models to automotive DL frameworks under a simulated temperature-varying environment. The total testing time for ThermalGuardian is 24 hours, with each temperature scenario in Table \ref{table_temperature_scenario} allocated 4 hours. For fairness, all baselines are also executed for 24 hours. During testing, ThermalGuardian records all model execution logs and output tensors, and detects automotive DL framework bugs as described in Section \ref{bug detection}.

\subsection{Research Questions}
\label{research questions}

Our evaluation focuses on four pivotal research questions as follows:

\begin{itemize}
    \item \textbf{RQ1 (Testing Effectiveness)}: How effectively does ThermalGuardian detect automotive DL framework bugs compared to baselines?
    
    \item \textbf{RQ2 (Bug Root Causes)}: What are root causes of detected DL framework bugs?
    
    \item \textbf{RQ3 (Efficiency \& Coverage)}: To what extent does ThermalGuardian outperform the baselines in terms of testing efficiency and coverage?
    
    \item \textbf{RQ4 (Ablation Study)}: How do key components (e.g., environment simulation and model mutation) impact ThermalGuardian's overall effectiveness?
\end{itemize}

In RQ1, we assess the bug detection effectiveness of ThermalGuardian against baselines by comparing the number of detected crashes, NaN bugs and heavy inconsistencies. In RQ2, we analyze the root causes of detected bugs for framework reliability. In RQ3, we measure testing efficiency (e.g., average bug detection time) and testing coverage (e.g., operator coverage) to quantify ThermalGuardian's improvements in efficiency and sufficiency. In RQ4, we conduct an ablation study to evaluate the individual contributions of ThermalGuardian's two key components: temperature-varying environment simulation and model mutation focusing on temperature-sensitive operators.

\subsection{RQ1: Testing Effectiveness}
\label{RQ1: testing effectiveness}
Following existing methods \cite{gandalf,muffin}, we focus on two types of bugs when evaluating test effectiveness: crashes and NaN bugs. Additionally, the number of detected heavy inconsistencies between cloud-end and automotive DL frameworks are evaluated, as these inconsistencies often indicate framework incompatibilities and may degrade performance when models are migrated across DL frameworks (see Section \ref{bug detection} for heavy inconsistencies' test oracle). For fairness, we execute ThermalGuardian and all baselines for 24 hours respectively. The workstation for executing all methods are kept the same. The experimental results are shown in Table \ref{table_total_effectiveness}. The first column of the table represents the method names, and next three columns respectively represent the number of crashes, NaN bugs, and heavy inconsistencies detected by each method. From this table, we can find that ThermalGuardian successfully detects 18 crashes, 3 NaN bugs, and 15 heavy inconsistencies in Apollo's PaddleInference and AutoWare's TensorRT. In contrast, LEMON does not trigger any bugs, Muffin only triggers 2 crashes and 2 heavy inconsistencies, and Gandalf only triggers 3 crashes. The improvement in the number of detected bugs shows ThermalGuardian's outstanding effectiveness. All detected bugs are reported to the public community, with 12 crashes and 3 NaN bugs confirmed by community developers. Other detected bugs are still waiting for developers' response.
% 这里把数据都填好，然后做好数据之间的分析比较。

\begin{table}[h]
\centering
\caption{Comparison in Testing Effectiveness}
\label{table_total_effectiveness}
\begin{tabular}{cccc}
\hline
Method & Crashes & NaNs & Inconsistencies \\
\hline
ThermalGuardian & 18 & 3 & 15 \\ 
LEMON & 0 & 0 & 0 \\ 
Muffin & 2 & 0 & 2 \\ 
Gandalf & 3 & 0 & 0 \\ \hline
\end{tabular}
\end{table}

Additionally, in Table \ref{table_temperature_scenario}, ThermalGuardian simulates six temperature-varying scenarios. The testing effectiveness of each scenario is compared in Table \ref{table_scenario_effectiveness}. The first column of the table represents the scenario names, and the next three columns respectively represent the number of crashes, NaN bugs, and heavy inconsistencies detected by ThermalGuardian in each scenario. The same bug may be detected across different scenarios. Crashes with identical execution logs and NaNs triggered by the same model are considered as redundant bugs. When summarizing the total number of bugs detected by ThermalGuardian across various scenarios, these redundant bugs are not counted repeatedly. The experimental results show that, in Scenario 1 and 4, ThermalGuardian detects the most DL framework quality issues. While in other four scenarios, ThermalGuardian detects relatively fewer quality issues. The difference among scenarios in testing effectiveness is due to the difference in the range of temperature fluctuations. Specifically, in Scenario 1 and 4, the GPU temperature varies between -40 °C and 90 °C. While the GPU temperature varies between -40 °C and 40 °C in Scenario 2 and 5, and varies between 40 °C and 90 °C in Scenario 3 and 6. The temperature achieves the highest fluctuation range (130°C) in Scenario 1 and 4, and Thermal Guardian triggers the most quality issues under these two scenarios. This results show that larger temperature range contributes to detecting more bugs.

\begin{table}[h] 
\centering
\caption{Testing Effectiveness in ThermalGuardian's Different Scenarios }
\label{table_scenario_effectiveness}
\begin{tabular}{m{8.25cm}<{\centering}ccc}
\hline
Scenario & Crashes & NaNs & Inconsistencies \\
\hline
Heavy Workloads in Cold Environment & 10 & 3 & 13 \\ 
Nominal Workloads in Cold Environment & 6 & 1 & 2 \\ 
Heavy Workloads in Nominal Environment & 5 & 1 & 3 \\ 
Cooling in Cold Environment After Heavy Workloads & 9 & 3 & 12 \\ 
Cooling in Cold Environment After Nominal Workloads & 4 & 0 & 1 \\ 
Cooling in Nominal Environment After Heavy Workloads & 6 & 1 & 2 \\
\hline
\end{tabular}
\end{table}

\begin{center}
\fcolorbox{black}{lightgray}{\parbox{.95\linewidth}{\textit{Answer to RQ1:} ThermalGuardian successfully detects 18 crashes and 3 NaN bugs, and 15 heavy inconsistencies in Apollo's PaddleInference and AutoWare's TensorRT, significantly surpassing all baselines (triggering 3 crashes and 2 inconsistencies at most). Additionally, ThermalGuardian detects more bugs in scenarios with larger temperature fluctuation range. 
}}
\end{center}

\subsection{RQ2: Root Causes}
\begin{figure}[htpb]
    \begin{subfigure}[htpb]{0.47\textwidth}  
        \centering
        \includegraphics[width=0.81\textwidth]{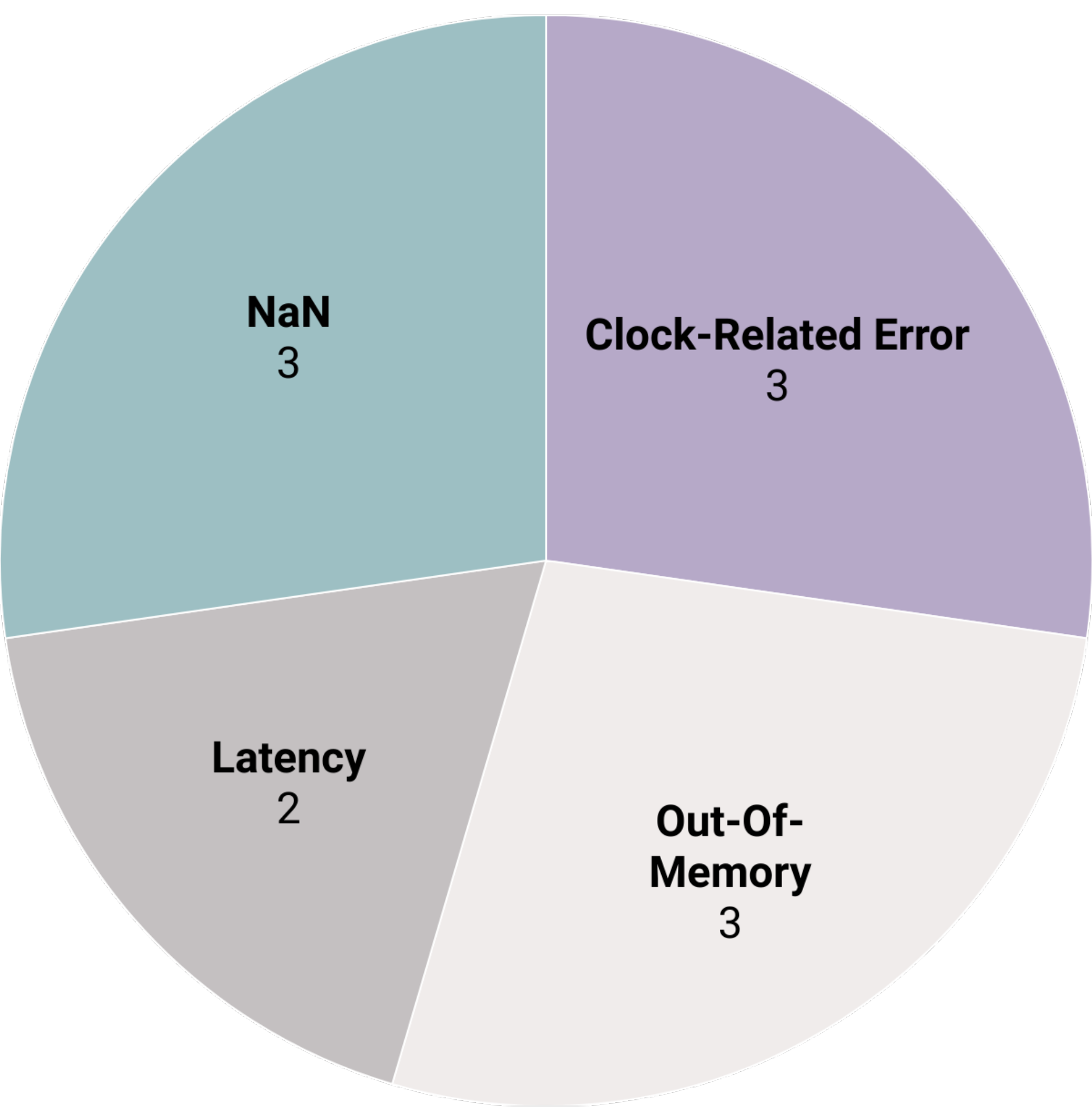}  % 宽度从0.9增加到0.95
        \caption{Root Causes of Temperature-Related Bugs}
        \label{figure_root_cause_temperature_related}
    \end{subfigure}
    \hfill
    % 右边图片（相应调整）
    \begin{subfigure}[htpb]{0.52\textwidth} 
        \centering
        \includegraphics[width=0.74\textwidth]{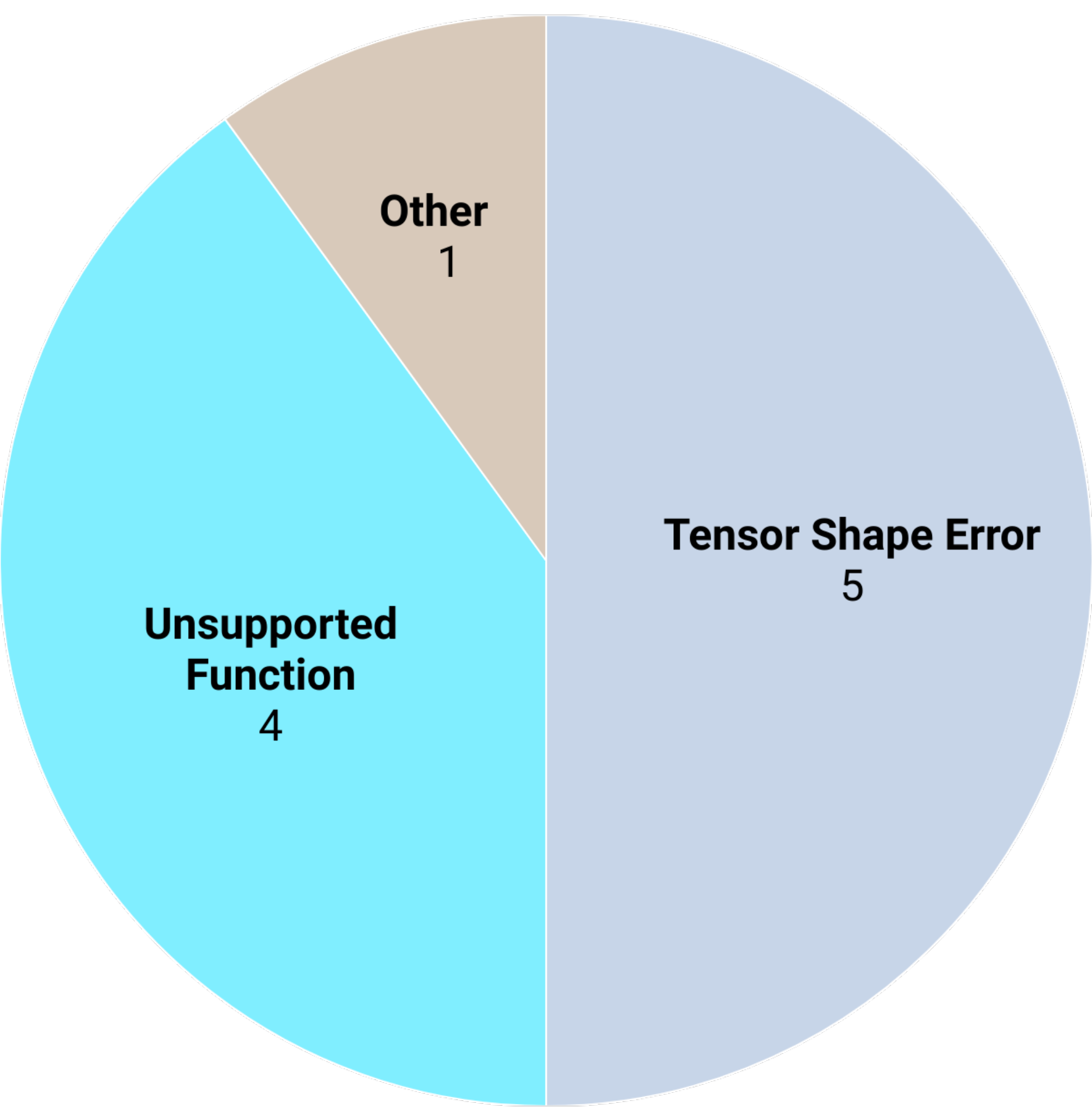}
        \caption{Root Causes of Non-Temperature-Related Bugs}
        \label{figure_root_cause_non_temperature_related}
    \end{subfigure}
    \caption{Root Causes of Detected Bugs}
\end{figure}

ThermalGuardian successfully detects 21 bugs, including 18 crashes and 3 NaN bugs. To analyze the root causes of detected crashes, two researchers with expertise in autonomous driving systems separately check the execution logs related to each bug, with any discrepancies resolved through formal arbitration by senior developers from the relevant open-source communities. Among detected bugs, 11 are related to temperature fluctuations, while 10 are not. The root causes of temperature-related and non-temperature-related bugs are shown in Figure \ref{figure_root_cause_temperature_related} and Figure \ref{figure_root_cause_non_temperature_related}.

\subsubsection{Root Causes of Temperature-Related Bugs}
Temperature-related bugs stem from four root causes: Latency, Clock-Related Error, NaN, and Out-Of-Memory.

\textbf{1) Latency (2/11, 18.18\%).} These bugs arise from latency issues caused by temperature-induced frequency decrease. Among them, one bug occurs in AutoWare's TensorRT when GPU frequency decrease prolongs execution time of time-series operators, causing delays that exceed the maximum system waiting time, thus leading to the GPU kernel being forcibly terminated due to a timeout error. Another bug occurs in Apollo's PaddleInference when GPU frequency decrease causes the latency in the initialization of the GPU driver, thus leading to GPU unresponsiveness.

\textbf{2) Clock-Related Error (3/11, 27.27\%).} These bugs arise from temperature-induced clock signal jitters. Among them, the first bug occurs in Apollo's PaddleInference when clock jitters disrupt the transmission or verification of GPU kernel launch parameters, and thus triggering an CUBLAS error. The second bug occurs in Apollo's PaddleInference when clock jitters disrupt gating signal synchronization, triggering an synchronization error. The last bug occurs in Autoware's TensorRT during compute-intensive operations like GEMM convolution in cuDNN, where clock jitters interfere with the synchronization process of cuDNN's internal state machine, causing critical steps to miss timing deadlines and triggering internal state inconsistency errors.

\textbf{3) NaN (3/11, 27.27\%).} These three bugs occur when output tensors contain unexpected NaN values. These NaNs originate from floating-point errors caused by temperature-induced GPU frequency decrease. Specifically, GPU frequency decrease causes floating-point operations to fail to complete all carry operations within the specified time, leading to data truncation that results in out-of-bound floating-point values or representation errors.

\textbf{4) Out-of-Memory (3/11, 27.27\%).} These three bugs originate from out-of-memory issues caused by insufficient GPU memory. One occurs in Apollo's PaddleInference, while the other two occur in AutoWare's TensorRT. For the same models under identical computational resources, these bugs do not occur in temperature-stable indoor environments but manifest in temperature-varying environments. This is because temperature fluctuations induce frequency variations, exacerbating memory fragmentation and ultimately leading to out-of-memory errors.

\subsubsection{Root Causes of Non-Temperature-Related Bugs}
Non-Temperature-Related bugs stem from three root causes: Unsupported Function, Tensor Shape Error, and Other.

\textbf{1) Unsupported Function (4/10, 40\%).} These bugs originate from unsupported functions in PaddleInference and TensorRT. Among them, in Apollo's PaddleInference, one bug stems from the lack of support for the API called $nms\_gpu$, and two bugs arise because PaddleInference does not support constructing a memory descriptor using a format tag. Another bug occurs in AutoWare's TensorRT due to invalid memory access.

\textbf{2) Tensor Shape Error (5/10, 50\%).} These bugs are caused by issues related to tensor shape. Among them, in Apollo's PaddleInference, three bugs occur because some time-series operators (e.g., RNN, LSTM, and GRU) do not support dynamic tensor shape, and one bug occurs because when the model contains these time-series operators, other operators that originally supported dynamic tensor shape lose this capability. In AutoWare's TensorRT, a bug occurs during inference optimization when not all tensor data is fully loaded into the GPU. This causes a mismatch between the operator's expected tensor shape and the actual tensor shape received.

\textbf{3) Other (1/10, 10\%).} An unexpected TensorRT setup failure is triggered. Due to the lack of more detailed descriptive information, researchers and community developers have not yet identified the cause of this failure and require further analysis and localization.

\begin{center}
\fcolorbox{black}{lightgray}{\parbox{.95\linewidth}{\textit{Answer to RQ2:} ThermalGuardian detects 11 temperature-related bugs and 10 non-temperature related bugs. The root causes of temperature-related bugs include: Latency, Clock-Related Error, NaN, and Out-Of-Memory. The root causes of non-temperature-related bugs include: Unsupported Function, Tensor Shape Error, and Other.
}}
\end{center}

\subsection{RQ3: Efficiency \& Test Coverage}
\label{RQ3: efficiency & test coverage}

\begin{figure}
    \centering
    \includegraphics[width=0.8\linewidth]{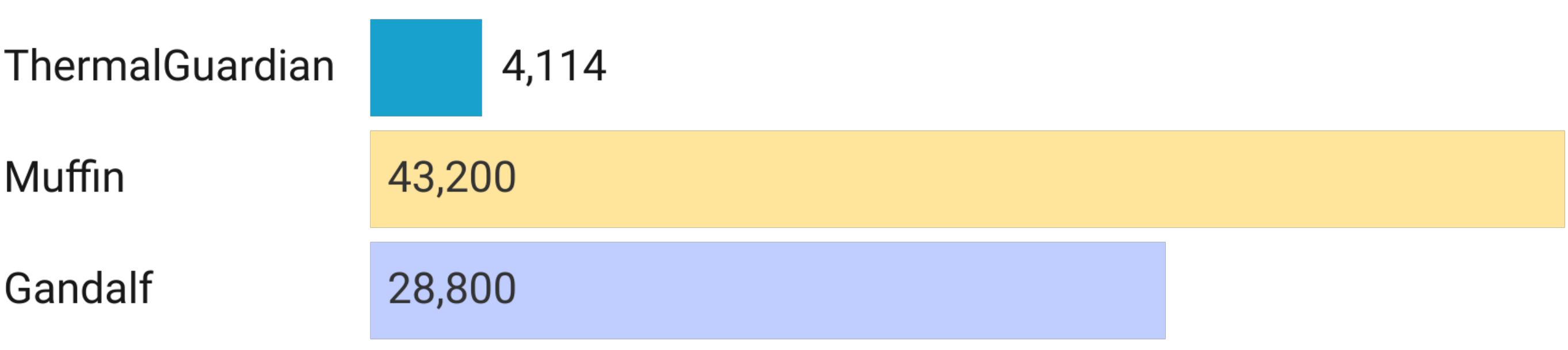}
    \caption{Average Time Consumption for Detecting Each Automotive DL Framework Bug (Seconds)
    }
    \label{table_efficiency}
\end{figure}

To evaluate ThermalGuardian's efficiency, we compare ThermalGuardian with all baselines on the average time consumption for detecting each automotive DL framework bug, as shown in Figure \ref{table_efficiency}. In this figure, the length of bars and the number on bars indicate the average time in seconds required by each method to detect one bug. Since LEMON does not trigger any bugs, we do not list it in the figure. From this figure, we can find that ThermalGuardian takes 4,114 seconds on average to detect each unique automotive DL framework bug, which is 7 times faster than Gandalf, highlighting the critical role of temperature-related design in testing automotive DL frameworks.

% \begin{table}[htpb]
%     \caption{Comparison in Efficiency}
%     \centering
%     \begin{tabular}{|c|c|}
%     \hline
%          \textbf{Method} & \textbf{Time Per Bug (s)} \\ \hline
%          ThermalGuardian  & 4114 \\ \hline
%          LEMON  & N/A \\ \hline
%          Muffin  & 43200 \\ \hline
%          Gandalf  & 28800 \\ \hline
%     \end{tabular}
%     \label{table_efficiency}
% \end{table}

To evaluate ThermalGuardian's testing sufficiency, we compare ThermalGuardian with all baselines on the testing coverage. Unlike conventional software systems that rely on the control flow, DL frameworks are data-driven. Consequently, conventional testing sufficiency metrics such as code coverage or branch coverage cannot properly evaluate DL framework testing. In line with prior work \cite{comet}, we adopt operator coverage (also referred to as API coverage) to assess the testing sufficiency of ThermalGuardian. This metric measures the proportion of operators invoked by generated test input models out of all operators supposed to be tested. Higher operator coverage indicates more sufficient testing. Furthermore, to illustrate the test sufficiency related to the temperature, we compare ThermalGuarian with all baselines on the coverage of temperature-sensitive operators. The experimental results are shown in Table \ref{table_coverage}. In this table, the first column represents the method name. The following two columns represent the operator coverage and the temperature-sensitive operator coverage, respectively. From this table, we can find that ThermalGuarian achieves the highest operator coverage (85\%) among all baselines, which shows ThermalGuardian's outstanding test sufficiency. Additionally, ThermalGuardian attains 100\% temperature-sensitive operator coverage, whereas LEMON, Muffin, and Gandalf achieve only 31.25\%, 25\%, and 25\%, respectively. These results indicate that existing methods fail to adequately account for the impact of temperature on the quality of automotive DL frameworks. The test input models generated by existing methods contain relatively few temperature-sensitive operators. In contrast, the model mutation rules in ThermalGuardian are beneficial to generating test input models that achieve significantly higher coverage of temperature-sensitive operators.

\begin{table}[htpb]
    \caption{Comparison in Testing Coverage}
    \centering
    \begin{tabular}{cm{4cm}<{\centering}m{5cm}<{\centering}}
    \hline
         \textbf{Method} & \textbf{Operator Coverage} & \textbf{Temperature-Sensitive Operator Coverage} \\ \hline
         ThermalGuardian & 85\% & 100\% \\  
         LEMON & 70\% & 31.25\% \\ 
         Muffin & 75\% & 25\% \\ 
         Gandalf & 80\% & 25\% \\ \hline
    \end{tabular}
    \label{table_coverage}
\end{table}

\begin{center}
\fcolorbox{black}{lightgray}{\parbox{.95\linewidth}{\textit{Answer to RQ3:} ThermalGuardian significantly outperforms all baselines in both testing efficiency and coverage. For efficiency, ThermalGuardian detects each unique automotive DL framework bug in 4,114 seconds--7 times faster than the best of existing methods (28,800 seconds). For coverage, ThermalGuardian achieves an 85\% operator coverage (surpassing all baselines). Additionally, ThermalGuardian achieves a 100\% temperature-sensitive operator coverage, while the best of all baselines attains only 31.25\%. These results confirm that existing methods fail to adequately consider temperature impacts due to insufficient inclusion of temperature-sensitive operators, whereas ThermalGuardian's model mutation rules effectively overcome this limitation.
}}
\end{center}

\subsection{RQ4: Ablation Study}
\label{RQ4: ablation study}

ThermalGuardian proposes two key innovations: the model mutation rules based on temperature-sensitive operators and the temperature-varying model deployment environment simulation. To quantify their contributions, we design three baselines:

\begin{itemize}
    \item $ThermalGuardian_{mutation}$: disables temperature-varying model deployment environment simulation, deploying models in indoor temperature-stable environments while retaining all model mutation rules.
    \item $ThermalGuardian_{simulation}$: disables model mutation rules based on temperature-sensitive operators. Generates test input models with simply non-temperature-sensitive operators and conduct testing within the simulated temperature-varying environment.
    \item $ThermalGuardian_{none}$: disables both two innovations, serving as a minimal baseline.
\end{itemize}

We execute each baseline for 24 hours respectively under the same workstation introduced in Section \ref{experimental setup}. The experimental results are shown in Table \ref{table_ablation study}. The first column of the table indicates the baseline names, and the next three columns respectively show the number of crashes, NaN bugs, and heavy inconsistencies detected by each baseline within 24 hours. As shown in the table, $ThermalGuardian_{simulation}$ (detecting 12 crashes, 2 NaN bugs, and 10 heavy inconsistencies) detects more quality issues than $ThermalGuardian_{none}$ (detecting only 2 crashes and 2 heavy inconsistencies), indicating that ThermalGuardian's temperature-varying model deployment environment simulation is effective. Furthermore, $ThermalGuardian$ (detecting 18 crashes, 3 NaN bugs, and 15 heavy inconsistencies) detects more quality issues than $ThermalGuardian_{simulation}$, demonstrating that ThermalGuardian's model mutation rules can further improve testing effectiveness in simulated environments with varying temperatures.

\begin{table}[htpb]

    \caption{Ablation Study on ThermalGuardian's Environment Simulation and Model Mutation}
    \centering
    \begin{tabular}{cccc}
    \hline
         \textbf{Baseline} & \textbf{Crashes} & \textbf{NaNs} & \textbf{Inconsitencies} \\ \hline
         $ThermalGuardian$ & 18 & 3 & 15\\ 
         $ThermalGuardian_{mutation}$ & 2 & 0 & 3 \\ 
         $ThermalGuardian_{simulation}$ & 12 & 2 & 10 \\ 
         $ThermalGuardian_{none}$ & 2 & 0 & 2\\ \hline
    \end{tabular}
    \label{table_ablation study}
\end{table}

\begin{center}
\fcolorbox{black}{lightgray}{\parbox{.95\linewidth}{\textit{Answer to RQ4:} The model mutation rules focusing on temperature-sensitive operators and the temperature-varying model deployment environment simulation both contribute to ThermalGuardian's testing effectiveness. Among them, the latter (environment simulation) independently enhances the testing effectiveness, while the former (mutation rules) further improves effectiveness on this basis.
}}
\end{center}
\section{Threats to Validity}
\label{threats to validity}
In this section, we discuss the threats to validity in our work, including internal validity, external validity, and construct validity.

Internal validity concerns the temperature simulation leveraging Newton's law of cooling. Newton's law of cooling is a typically simplification of complex temperature dynamics, and we have carefully calibrated our cooling coefficient parameters according to NVIDIA's official website. In this way, our simulation properly replicates realistic GPU temperature fluctuations.

External validity concerns the generalizability of our findings. ThermalGuardian tests on two widely-used industry-leading automotive DL frameworks—PaddleInference in Apollo and TensorRT in AutoWare. By making ThermalGuardian fully open-source with detailed documentation and modular architecture, we enable the research community to easily extend testing to additional DL frameworks, thereby significantly expanding the scope of our method and ensuring broad applicability of our findings across the autonomous driving ecosystem.

Construct validity concerns the design of model mutation rules and DVFS emulation. Our model mutation rules target three types of temperature-sensitive operators, but could potentially have missed other types. We have proactively addressed this concern through our Mutation Rule 8 (Random Operator Replacement) that systematically explores the combination among other operators. Additionally, we conduct our DVFS emulation according to existing works related to DVFS, and we have extensively calibrated the low-temperature scaling coefficient $\gamma$ and the hight-temperature throttling coefficient $\alpha$ according to NVIDIA's official data, effectively eliminating concerns about model fidelity and ensuring that our experimental results faithfully represent actual framework performance in temperature-varying environments.
\section{Related Works}
\label{related works}

\subsection{Deep Learning Framework Testing}
\label{deep learning framework testing}
The qualities issues within DL frameworks have attracted substantial research attention in recent years \cite{DLbug1,DLbug2,DLbug3,DLbug4,DLbug5,DLbug6,DLbug7,DLbug8}. Many testing methods are proposed to ensure the quality of DL frameworks. These methods are categorized into interface-based testing that generates tensors to invoke individual DL framework operators and model-based testing that generates both test input models and corresponding tensors. Within interface-based testing, Predoo \cite{predoo} designs tensor mutation rules to detect precision errors, FreeFuzz \cite{freefuzz} extracts insights from real-world code executions for API-level fuzzing, DeepREL \cite{DeepREL} infers API relationships for test generation, Duo \cite{duo} applies mutation rules with genetic algorithm guidance, and EAGLE \cite{eagle} systematically invokes equivalent operators. Within model-based testing, Cradle \cite{cradle} utilizes publicly available models as test inputs. LEMON \cite{lemon} mutates seed models through designed mutation rules. Gandalf \cite{gandalf} employs context-free grammar with deep-Q networks to generate test input models. Muffin \cite{muffin} firstly designs the connection between operators connections within each test input model, and then specifies each operator type to generate new models. DLLEN \cite{dllen} and TitanFuzz \cite{titanfuzz} leverage large language models for test input model generation. However, these existing DL framework testing methods simply explore the combination between DL framework operators, fail to consider the effect of temperature fluctuations. To bridge this gap, in our method, ThermalGuardian simulates a temperature-varying model deployment environment, and designs model mutation rules focusing on temperature-sensitive operators.

\subsection{DVFS in GPU}
\label{dvfs in gpu}

Dynamic Voltage and Frequency Scaling (DVFS) serves as a fundamental mechanism for frequency management in GPU computing systems. Existing research has established critical foundations for understanding DVFS mechanism. Mei et al. \cite{dvfs_related_work_1} systematically characterize voltage and frequency scaling effects on NVIDIA architectures, providing empirical data on power-performance tradeoffs. Tang et al. \cite{dvfs_related_work_2} investigate frequency scaling for deep learning workloads, quantifying the relationship between computational throughput and power consumption across different GPU generations. Karzhaubayeva et al. \cite{dvfs_related_work_8} develop coordinated CPU-GPU frequency control strategies, demonstrating effective energy-delay product optimization through integrated governance. Nabavinejad et al. \cite{dvfs_related_work_4} introduce the BatchDVFS approach that dynamically coordinates batching and frequency selection to maintain performance under power constraints. Guerreiro et al. \cite{dvfs_related_work_5} create detailed power models that decompose GPU energy consumption into constituent components, enabling precise DVFS impact prediction. Han et al. \cite{dvfs_related_work_6} formulate latency models that capture non-linear relationships between frequency scaling and inference performance. Nabavinejad et al. \cite{dvfs_related_work_7} propose the PIT framework that combines precision selection with frequency scaling for enhanced efficiency. These studies collectively provide the theoretical foundation for ThermalGuardian's approach to simulating frequency variations in response to temperature fluctuations during the automotive DL framework testing.

\section{Conclusion}
\label{conclusion}

In this work, we propose ThermalGuardian, an automotive DL framework testing method that simulates GPU temperature fluctuations using Newton's law of cooling and dynamically controls GPU frequency via temperature-aware DVFS mechanisms. By designing eight model mutation rules targeting temperature-sensitive operators (e.g., compute-intensive GEMM convolutions, high/mixed-precision operators, and time-series RNN/LSTM/GRU units), ThermalGuardian generates diverse test input models capable of exposing temperature-related automotive DL framework bugs. The evaluation is conducted on Apollo's PaddleInference and AutoWare's TensorRT. The experimental results show that ThermalGuardian successfully detects 11 temperature-related bugs and 10 non-temperature-related bugs, which is the most among all baselines. Additionally, ThermalGuardian is 7 times faster than the best baseline, and achieves the highest operator coverage (85\%) and temperature-sensitive operator coverage (100\%). These experimental results validate ThermalGuardian's effectiveness in detecting automotive DL framework bugs, underscoring its role in enhancing the robustness of autonomous driving systems. Future works will extend temperature simulation to multi-GPU configurations and refine mutation rules for emerging operator types.

\section{Data Availability}

We make all of data and source code publicly available on Github:  \url{https://github.com/ThermalGuardian/ThermalGuardian}.

\bibliographystyle{ACM-Reference-Format}

\bibliography{sample-base}
\end{document}